\documentclass{article}

\usepackage{arxiv}

\usepackage[utf8]{inputenc} 
\usepackage[T1]{fontenc}    
\usepackage{hyperref}       
\usepackage{url}            
\usepackage{booktabs}       
\usepackage{amsfonts}       
\usepackage{nicefrac}       
\usepackage{microtype}      
\usepackage{lipsum}		
\usepackage{graphicx}
\usepackage{doi}
\usepackage[round]{natbib}   
\usepackage{amsmath}
\usepackage{mathrsfs}
\usepackage{amssymb}
\usepackage{svg}
\usepackage{graphicx}
\usepackage{float}

\usepackage{color} 
\usepackage{todonotes} 

\title{Soft $Q(\lambda)$: A multi-step off-policy method for entropy regularised reinforcement learning  using eligibility traces}

\author{Pranav Mahajan  \\
	University of Oxford\\
	\texttt{pranav.mahajan@ndcn.ox.ac.uk} \\
	\And
    Ben Seymour  \\
	University of Oxford\\
    \texttt{ben.seymour@ndcn.ox.ac.uk} \\
}

\renewcommand{\shorttitle}{Soft $Q(\lambda)$}

\begin{document}
\maketitle


\begin{abstract}
Soft Q-learning has emerged as a versatile model-free method for entropy-regularised reinforcement learning, optimising for returns augmented with a penalty on the divergence from a reference policy. Despite its success, the multi-step extensions of soft Q-learning remain relatively unexplored and limited to on-policy action sampling under the Boltzmann policy. In this brief research note, we first present a formal $n$-step formulation for soft Q-learning and then extend this framework to the fully off-policy case by introducing a novel Soft Tree Backup operator. Finally, we unify these developments into Soft $Q(\lambda)$, an elegant online, off-policy, eligibility trace framework that allows for efficient credit assignment under arbitrary behaviour policies. Our derivations propose a model-free method for learning entropy-regularised value functions that can be utilised in future empirical experiments.
\end{abstract}


\section{Introduction}
Entropy-regularised reinforcement learning (RL) improves exploration, robustness and stability during learning by augmenting the reward objective with a penalty on the divergence from a reference policy (or a default policy) \citep{haarnoja2017reinforcement, van2019composing}. Entropy-regularised RL has its roots in Linear MDPs proposed by \citep{kappen2005linear, todorov2006linearly, todorov2009efficient}, sharing the same objective function. When the reference policy is set to a uniform random policy, it reduces to the special case of max entropy RL \citep{ziebart2010modeling}. One of the core features of such methods is that instead of learning a single deterministic behaviour that has the highest cumulative reward, the resulting policies try to learn all of the ways of performing the task, explicitly maximising the entropy of the corresponding policy. Such a stochastic policy is optimal when we consider the connection between optimal control and probabilistic inference \citep{todorov2008general}.

A number of methods have been proposed, including Z-learning \citep{todorov2006linearly, todorov2009efficient}, maximum entropy inverse
RL \citep{ziebart2008maximum}, approximate inference using message passing \citep{toussaint2009robot}, $\Psi$-learning \citep{rawlik2012stochastic}, and G-learning \citep{fox2015taming}, as well as more
recent proposals in deep RL such as PGQ \citep{o2016combining}. Previous work has also established equivalence between policy gradient methods and soft Q-learning, where the optimal policy is shown to be a Boltzmann distribution of the action-values \citep{schulman2017equivalence}. However, extending this framework to multi-step estimation introduces significant limitations. Specifically, the $n$-step soft Q-learning estimator proposed by \citet{schulman2017equivalence} is unbiased only when trajectories are sampled using the target (soft-optimal) Boltzmann policy. This on-policy constraint restricts the algorithm's utility in fully off-policy regimes or settings with arbitrary exploration schedules.

In this research note, we bridge this gap by extending soft Q-learning to a fully off-policy, multi-step regime. We first formalise a stepwise $n$-step soft Q-learning formulation. While off-policy learning can be achieved via importance sampling, it suffers from high variance. Therefore, we extend this framework by introducing a novel Soft Tree Backup operator that leverages the recursive relationship between the state-value function $V_Q$ and the action-value function $Q$. This operator handles entropy terms over multiple time steps without requiring knowledge of the behaviour policy, effectively eliminating the on-policy bias inherent in standard $n$-step soft backups. Finally, we unify these developments into Soft $Q(\lambda)$, an elegant eligibility trace framework that can enable efficient, online, off-policy credit assignment. Our derivations demonstrate that entropy-regularised value functions can be learned stably under arbitrary behaviour policies without the reliance on target networks or fixed exploration schedules, providing a theoretically grounded toolkit for robust reinforcement learning.

\section{Background}

\subsection{Reinforcement learning in MDPs}
Let the environment be a Markov Decision Process, where at time $t = 0,1,2,...$, the agent is in state $s_t \in \mathcal{S}$ and takes action $a_t \in \mathcal{A}$ and receives the next state $s_{t+1} \in \mathcal{S}$ and the reward $r_{t+1} = r(s_t,a_t) \in \mathcal{R}$ giving rise to trajectories $s_0, a_0, r_1, s_1, a_1, r_2, ...$. The dynamics of MDP are given by the conditional probability $p(s', r | s, a) \doteq \mathrm{Pr}(s_t = s', r_t = r | s_{t-1} = s, a_{t-1} = a)$.

The discounted return at time $t$ is given by $G_t = r_{t+1} + \gamma r_{t+2} + \gamma^2 r_{t+3} + ... = \sum_{k=0}^{\infty} \gamma^{k} r_{t+k+1}$, where $\gamma \in [0,1]$. Policy $\pi (a|s)$ is a mapping from states to the probabilities of choosing each possible action. The value function of a state $s$ under the policy $\pi$ is is the expected return when starting in $s$ and following $\pi$ thereafter, which is formalized as $V_{\pi} \doteq \mathbb{E}_{\pi} [G_t | s_t=s], \forall s \in \mathcal{S}$. Similarly, the value of taking action $a$ in state $s$ and following policy $\pi$ thereafter is given by the Q-value or the action-value function, $Q_{\pi}(s,a) \doteq  \mathbb{E}_{\pi} [G_t | s_t=s, a_t=a]$.

The Bellman equation of a value function $v_\pi$ is a fundamental property in reinforcement learning expressing the recursive relationship between the value of a state and the value of its possible successor states. 

\begin{equation}
    V_\pi(s) \doteq \mathbb{E}_{a \sim \pi(\cdot | s)}\mathbb{E}_{ (s',r) \sim p(s',r|s,a)} [r+\gamma V_\pi(s')], \forall s \in \mathcal{S}
    \label{eqn:RL_Bellman_equation}
\end{equation}

Since value functions define a partial ordering over policies, there exists at least one optimal policy $\pi^*$ that is better than all policies, where a policy $\pi \geq \pi'$ if and only if $V_\pi(s) \geq V_{\pi'}(s), \forall s \in \mathcal{S}$. The optimal state-value function is $V^{*}(s) \doteq \max_{\pi} V_\pi(s), \forall s \in \mathcal{S}$. Similarly, the optimal action-value function is $Q^{*}(s, a) \doteq \max_{\pi} Q_\pi(s,a) = \mathbb{E}[r_{t+1} + V^{*}(s_{t+1})| s_t=s, a_t=a]$. Once we have the optimal action-values, one can simply perform actions greedily to get the optimal policy $\pi^{*} = [\mathcal{G}Q^*](s) = \arg \max_{a} Q^*(s,a)$.

The recursive Bellman equations can also be written for the value function under the optimal policy, referred to as the Bellman optimality equations:

\begin{equation}
     V^*(s) = \max_a\mathbb{E}_{ (s',r) \sim p(s',r|s,a)}[r+\gamma V^*(s')]
    \label{eqn:RL_bellmanoptimalityoperator}
\end{equation}

\subsection{Entropy-regularised reinforcement learning in Linear MDPs}
\label{sec:ent-RL}
Entropy-regularised RL \citep{todorov2006linearly, todorov2009efficient, van2019composing} augments the reward function with a term that penalises deviating from some default policy $\pi^d$, essentially making “soft” assumptions about the future policy (in the form of a stochastic action distribution). When $\pi^d$ is a uniform policy, this reduces to max entropy reinforcement learning \citep{ziebart2010modeling, haarnoja2017reinforcement}. The expected reward on taking action $a_t$ in state $s_t$ is given by $\mathbb{E}_{a_t\sim\pi} [r(s_t, a_t) - \tau D_{\mathrm{KL}}(\pi(\cdot|s_t) \| \pi^d(\cdot|s_t))]$, which can be further compactly written as $\mathbb{E}_{a_t\sim\pi} [r_{t+1} - \tau \mathrm{KL}(s_t)]$. Here, $\tau$ is the scalar temperature parameter, and $\mathrm{KL}(s_t)$ is the Kullback-Leibler divergence between the current policy $\pi$ and a default policy $\pi^d$ in state $s_t$. Thus, the entropy-augmented return is $G_t = 
\sum_{k=0}^{\infty} \gamma^{k} (r_{t+k+1} - \tau \mathrm{KL}(s_{t+k}))$.

The value function definitions under a policy $\pi$ at any timestep $t$ based on the entropy-augmented returns are as follows,

\begin{equation}
    V_{\pi}(s) \doteq  \mathbb{E}_{\pi} [G_t | s_t=s] = \mathbb{E}_\pi \left[ \sum_{k=0}^{\infty} \gamma^{k} (r_{t+k+1} - \tau \mathrm{KL}(s_{t+k})) \bigg| s_t=s\right]
\end{equation}

\begin{equation}
    Q_{\pi}(s,a) \doteq  \mathbb{E}_{\pi} [G_t | s_t=s, a_t=a] = \mathbb{E}_\pi \left[ r_{t+1} +  \sum_{k=1}^{\infty} \gamma^{k} (r_{t+k+1} - \tau \mathrm{KL}(s_{t+k})) \bigg| s_t=s, a_t=a \right]
\end{equation}

Note that this Q-function does not include the first KL penalty term ($\mathrm{KL}(s_t)$), as it does not depend on action $a_t$ which has already been chosen \citep{ziebart2010modeling, haarnoja2017reinforcement, schulman2017equivalence}. This gives the following relationship which holds for all policies $\pi$.

\begin{equation}
    V_{\pi}(s) =  \mathbb{E}_{a\sim\pi} [Q_{\pi}(s,a)] - \tau \mathrm{KL}(s)
    \label{eqn:V_pi_relationship}
\end{equation}


The Bellman equation is as follows:
\begin{equation}
    V_\pi(s) \doteq \mathbb{E}_{a \sim \pi(\cdot|s)}\mathbb{E}_{(s',r) \sim p(s',r|s,a)} [r - \tau \mathrm{KL}(s) +\gamma V_\pi(s')]
\end{equation}


Note, unlike the greedy (deterministic) policy [$\mathcal{G}Q](s) = \arg \max_a Q(s,a)$ in standard RL, the greedy (stochastic) policy in entropy-regularised RL is the Boltzmann policy ($\pi^{\mathcal{B}}_Q$).

\begin{equation}
    \pi^{\mathcal{B}}_Q(\cdot| s) = [\mathcal{G}Q](s) = \frac{\pi^d(a|s) \exp(Q(s,a)/\tau)}{\sum_{\mathcal{A}} \exp(Q_\pi(s,a')/\tau) \pi^{d}(a'|s)} 
    \label{eqn:boltzmannpolicy}
\end{equation}

Prior work \citep{todorov2006linearly, todorov2009efficient, haarnoja2017reinforcement, van2019composing} shows that this Boltzmann policy holds the two properties: (1) it is the optimal policy ($\pi^*=\pi^{\mathcal{B}}_{Q^*}$ ) i.e. it uniquely solves the Bellman optimality equations and (2) under the Boltzmann policy, the Bellman equation is equivalent to the "soft" Bellman equation, thus the value function $V_{\pi^{\mathcal{B}}_Q}(s) =V_Q(s)$, essentially performing a soft maximum operation over Q-values. 

\begin{equation}
\begin{split}
    V_Q(s) &= \tau \log \mathbb{E}_{a\sim\pi^d} \exp(Q_\pi(s,a)/\tau)\\ 
    &= \tau \log \sum_{\mathcal{A}} \exp(Q_\pi(s,a)/\tau) \pi^{d}(a|s) 
\end{split}
    \label{eqn:V_Q}
\end{equation}

Note, this log-sum-exp performs a soft maximum because, $\max\{x_1,..., x_n\} \leq \mathrm{softmax}(x_1,..., x_n) \leq \max\{x_1,..., x_n\} + \log(n)$.

\subsection{Off-policy model-free learning algorithms in Linear MDPs}
\label{sec:methods_algorithms}
Model-free algorithms do not assume a probabilistic model about state transitions and rewards but instead learn value functions through reward prediction errors. Here, we focus on online algorithms, such as soft Q-learning \citep{haarnoja2017reinforcement}, which update values continuously during episodes rather than waiting until the end, unlike offline algorithms like Z-learning \citep{todorov2006linearly}, a Monte Carlo control algorithm. We further particularly focus on off-policy algorithms like Soft Q-learning and our subsequent extensions.  

\textbf{Soft Q-learning (One-Step)}

We adopt soft Q-learning and extend it from the maximum entropy formulation to a relative entropy formulation. The Q-value update equation is given by:

\begin{equation}
    Q(s_t,a_t) \leftarrow  Q(s_t,a_t) + \alpha \delta_t,
\end{equation}

where $\alpha$ is the learning rate, and $\delta_t$ is the reward prediction error at timestep $t$, defined as:

\begin{equation}
    \delta_t = r_{t+1} + \gamma V_Q(s_{t+1}) - Q(s_t, a_t),
    \label{eqn:softQ_delta}
\end{equation}

where $V_Q$ is given by equation \ref{eqn:V_Q}. Deep RL implementations inspired by \citet{mnih2015human} may use a separate target network (e.g., $\underline{Q}$, resulting in $V_{\underline{Q}}$) to construct the loss function, which we exclude here for simplicity.

\section{Results: Multi-step Soft Q-learning}

This section presents novel update rules for multi-step extensions of soft Q-learning, where the agent learns from multiple steps rather than the most immediate step. Under the assumption that the state action values are approximately unchanging \citep{sutton2018reinforcement}, we can write the update rule for the N-step soft Q learning and its extension with eligibility traces, soft Q($\lambda$) using TD-errors. 

When following the Boltzmann policy, the N-step soft Q-learning is simply, 

\begin{equation}
    Q_{t+n}(s_t, a_t) \leftarrow Q_{t+n-1}(s_t, a_t) + \alpha \left(\sum_{k=t}^{min(T-1, t+n-1)} \gamma^{k-t} \delta_{k} \right).
\end{equation}

Where $T$ is the time step at which the episode terminated, $Q_{t+n}$ denotes the Q-value accessed or updated at timestep $t+n$, and the TD-errors are defined as follows. Because on-policy N-step soft Q-learning expands the expectation over the state-value function, the TD-error for $k>t$ must explicitly include the KL divergence penalty. This formulation mirrors the exact temporal difference error derived in Equation 43 of \citet{schulman2017equivalence} for their soft TD($\lambda$) algorithm:

\begin{equation}
    \delta_{k} = r_{k+1} - \tau \text{KL}_{k} + \gamma V_Q(s_{k+1}) - V_Q(s_{k})
    \label{eqn:softQ_delta_extension}
\end{equation}
(For $k=t$, the TD-error is the standard one-step error given by equation \ref{eqn:softQ_delta}).

However, if one is following a behavioural policy that is not the Boltzmann policy (equation \ref{eqn:boltzmannpolicy}), then we need a truly off-policy update rule. If the agent has access to the behavioural policy, then it can use importance sampling (detailed derivation provided in Appendix 1). However, this can lead to higher variance in the updates and requires access to the behavioural policy. Therefore, we derive an alternative method using a novel Soft Tree Backup operator, which does not require explicit knowledge about the behavioural policy. 

Crucially, because Tree Backup analytically computes the expectation over actions at each step, it reconstructs the soft value function internally. As shown in Appendix 1, this causes the explicit $\text{KL}$ penalties to perfectly cancel out, shifting the telescoping sum from state-values to action-values. This marks a fundamental departure from the on-policy formulation in \citet{schulman2017equivalence}: by mathematically absorbing the KL penalty, we establish a clean, action-value based TD-error that enables multi-step credit assignment regardless of the behaviour policy. The redefined Tree Backup TD-error for all steps $k \geq t$ becomes:

\begin{equation}
    \delta^{TB}_k = r_{k+1} + \gamma V_Q(s_{k+1}) - Q_{k-1}(s_k, a_k)
\end{equation}

The off-policy $n$-step update rule is thus:

\begin{equation}
    Q_{t+n}(s_t, a_t) \leftarrow Q_{t+n-1}(s_t, a_t) + \alpha \left(\sum_{k=t}^{min(T-1, t+n-1)} \delta^{TB}_k \prod_{i=t+1}^{k} \gamma \pi^{\mathcal{B}}_Q(a_i|s_i)  \right).
\end{equation}

We next extend these methods to incorporate eligibility traces. Under the Boltzmann policy, the Q-value update rule is, 

\begin{equation}
    Q_{t+1}(s,a) \leftarrow Q_{t}(s,a) + \alpha \delta_t e_t (s,a) \; \; \forall s,a
\end{equation}

and eligibility traces are updated as follows (in the tabular setting),

\begin{equation}
    e_t(s, a) = 
    \begin{cases}
        \gamma \lambda e_{t-1}(s, a) + 1, & \text{if } (s, a) = (s_t, a_t), \\
        \gamma \lambda e_{t-1}(s, a), & \text{otherwise},
    \end{cases}
\end{equation}

For this on-policy formulation, the TD-error ($\delta_t$) uses the state-value formulation from equation \ref{eqn:softQ_delta_extension} (substituting $k$ with $t$). Note, this algorithm is entirely online.

For a full off-policy Soft Q($\lambda$), we build upon the Tree Backup approach. The Q-value update rule remains the same, but it utilises the action-value TD-error ($\delta^{TB}_t$) which drops the KL penalty. Furthermore, the eligibility trace updates are adjusted to include the target policy $\pi^{\mathcal{B}}_Q$,

\begin{equation}
    e_t(s, a) = 
    \begin{cases}
        \gamma \lambda \pi^{\mathcal{B}}_Q (a_t|s_t)  e_{t-1}(s, a) + 1, & \text{if } (s, a) = (s_t, a_t), \\
        \gamma \lambda \pi^{\mathcal{B}}_Q (a_t|s_t) e_{t-1}(s, a), & \text{otherwise},
    \end{cases}
\end{equation}

All detailed derivations for N-step soft Q-learning and Soft Q($\lambda$) are provided in Appendices 1 and 2, respectively.

\section{Conclusions and a Neuroscientific Epilogue}

This note extends soft Q-learning to a multi-step, off-policy regime. By introducing a novel Soft Tree Backup operator and then extending to the Soft Q($\lambda$) framework, the method overcomes prior on-policy limitations and enables multi-step credit assignment under arbitrary, unknown behaviour policies.

This work, laying the theoretical foundations, is particularly useful to the neuroscience of learning and decision making. Recent work \citep{mahajan2025composing} utilises the benefits of entropy-regularised RL, such as optimal composition of multiple values and stable learning due to KL-regularisation, to provide a new theoretical update to the seminal theory of phasic dopamine responses \citep{schultz1997neural}. The proposed theory \citep{mahajan2026entropy} also attempts to unify several disparate heterogeneities between and within different dopamine targets, including recently observed action prediction errors \citep{greenstreet2025dopaminergic}, into the temporal difference RL framework. Meanwhile, model-based solutions to Linear MDPs have also been recently used to explain phenomena in human planning, grid fields, cognitive control and medial entorhinal cortex representations \citep{piray2021linear, piray2024reconciling}. Ultimately, these theoretical derivations provide a robust, model-free toolkit for entropy-regularised reinforcement learning, establishing a mathematical foundation for future empirical evaluations in complex environments.

\section*{Author Contributions}
PM: Conceptualisation, Formal Analysis, Writing – Original Draft Preparation, Writing – Review \& Editing. BS: Funding Acquisition, Supervision.
\section*{Acknowledgments}

Authors thank the funders: Wellcome Trust (214251/Z/18/Z, 203139/Z/16/Z and 203139/A/16/Z), IITP (MSIT 2019-0-01371) and JSPS (22H04998). This research was also partly supported by the NIHR Oxford Health Biomedical Research Centre (NIHR203316).  The views expressed are those of the author(s) and not necessarily those of the NIHR or the Department of Health and Social Care. For the purpose of open access, the authors have applied a CC BY public copyright licence to any Author Accepted Manuscript version arising from this submission.

\bibliography{references}

\newpage

\section*{Appendix}

\subsection*{Appendix 1: Novel derivations extending Soft Q-learning to N-step soft Q-learning}

In this section, we provide a detailed derivation of how soft Q-learning can be extended to N-step soft Q-learning. We will first begin with the on-policy setting, under the special case of Boltzmann policy (the stochastic optimal policy) and then extend it to a fully off-policy algorithm.

\textbf{N-step Soft Q-learning (on-policy with Boltzmann policy)}

N-step soft Q-learning incorporates multiple future rewards and KL penalties for deviating from the default policy, starting from the second time step onward. 

The N-step return at time $t$, after taking an action $a_t$ in state $s_t$ is defined as:

\begin{equation}
    G_{t:t+n} \doteq r_{t+1} + \gamma (r_{t+2} - \tau \text{KL}_{t+1}) + \gamma^2 (r_{t+3} - \tau \text{KL}_{t+2}) + \ldots + \gamma^{n-1} (r_{t+n} - \tau \text{KL}_{t+n-1}) + \gamma^n V_Q(s_{t+n}),
    \label{eqn: n-step return}
\end{equation}

Note that the KL penalty terms appear only from the second timestep onward, as the cost of deviating from the default policy affects subsequent actions. If the episode terminates at timestep $T$, which can be less than $t+n$, then we will see next that the summation of TD-errors is appropriately truncated to $min(T-1, t+n-1)$. 

We can rewrite $G_{t:t+n}$ in terms of the temporal difference (TD) error $\delta$, by adding and subtracting $\gamma V_Q(s_{t+1})$, $\gamma^2 V_Q(s_{t+2})$, $\gamma^3 V_Q(s_{t+3})$ and so on:

\begin{equation}
\begin{split}
    G_{t:t+n} &= (r_{t+1} + \gamma V_Q(s_{t+1}))+ \gamma (r_{t+2} - \tau \text{KL}_{t+1} + \gamma V_Q(s_{t+2}) -  V_Q(s_{t+1})) \\
    &\quad + \ldots + \gamma^{n-1} (r_{t+n} - \tau \text{KL}_{t+n-1} + \gamma V_Q(s_{t+n}) -V_Q(s_{t+n-1})).
\end{split}
\end{equation}

Simplifying, we obtain:

\begin{equation}
\begin{split}
    G_{t:t+n} &= (r_{t+1} + \gamma V_Q(s_{t+1})) + \sum_{k=t+1}^{min(T-1, t+n-1)} \gamma^{k-t} \delta_{k} \\
    &= Q_{t-1}(s_t,a_t) + (r_{t+1} + \gamma V_Q(s_{t+1}) - (Q_{t-1}(s_t,a_t)) + \sum_{k=t+1}^{min(T-1, t+n-1)} \gamma^{k-t} \delta_{k} \\
    &= Q_{t-1}(s_t,a_t) + \sum_{k=t}^{min(T-1, t+n-1)} \gamma^{k-t} \delta_{k} \\
\end{split}
\end{equation}

where the TD error $\delta_{k}$ at each timestep is given as follows.

If $k=t$, the same as soft Q-learning: 
\begin{equation}
    \delta_{t} = r_{t+1} + \gamma V_Q(s_{t+1}) - Q(s_{t}, a_{t})
    \label{eqn:delta_eq1}
\end{equation}

For $k\geq t$,
\begin{equation}
    \delta_{k} = r_{k+1} - \tau \text{KL}_{k} + \gamma V_Q(s_{k+1}) - V_Q(s_{k})
    \label{eqn:delta_eq2}
\end{equation}

The first TD error term, $\delta_t = r_{t+1} + \gamma V_Q(s_{t+1}) - Q(s_{t}, a_{t})$, does not include the KL penalty since it doesn't depend on the action $a_t$ which has already been chosen \citep{ziebart2010modeling, haarnoja2017reinforcement, schulman2017equivalence}.

Thus, the N-step soft Q-learning update rule is defined as:

\begin{equation}
    Q_{t+n}(s_t, a_t) \leftarrow Q_{t+n-1}(s_t, a_t) + \alpha \left(G_{t:t+n} - Q_{t+n-1}(s_t, a_t)\right),
\end{equation}

where $\alpha$ is the learning rate. The subscripts denote the timestep in the episode when the Q-value was used or updated. Note that n-step returns for $n > 1$ involve future rewards and states that are not
available at the time of transition from $t$ to $t + 1$. Thus, the first Q-update of state $s_t$ is performed at timestep $t+n$ and not $t$.

If the approximate action-values are unchanging, i.e. $Q_{t-1}(s_t, a_t) \simeq Q_{t+n-1}(s_t, a_t)$ (similar to Exercise 7.11 in \citet{sutton2018reinforcement}), then we can substitute the expression for $G_{t:t+n}$ to get:

\begin{equation}
    Q_{t+n}(s_t, a_t) \leftarrow Q_{t+n-1}(s_t, a_t) + \alpha \left(\sum_{k=t}^{min(T-1, t+n-1)} \gamma^{k-t} \delta_{k} \right).
\end{equation}

If the approximate action values are changing, then we will have an additional term of $Q_{t-1}(s_t, a_t) - Q_{t+n-1}(s_t, a_t)$ in the update.

\textbf{N-step Soft Q-learning (off-policy with importance sampling)}

We can now extend this to an off-policy algorithm that learns the Boltzmann policy ($\pi^{\mathcal{B}}_Q$) as the target policy while collecting data under any behavioural policy $b$. Considering that soft Q-learning is akin to expected SARSA for relative-entropy regularised objective, this derivation is similar to the N-step expected SARSA derivation \citep{sutton2018reinforcement}.

We define the importance sampling ratio as follows ($T$ is the last time step of the episode),

\begin{equation}
    \rho_{t:h} = \prod_{k=t}^{min(h, T-1)} \frac{\pi^{\mathcal{B}}_Q (a_k|s_k)}{b(a_k|s_k)}
\end{equation}

Now the update from the previous subsection can be replaced with its off-policy form, 

\begin{equation}
    Q_{t+n}(s_t, a_t) \leftarrow Q_{t+n-1}(s_t, a_t) + \alpha  \rho_{t+1:t+n-1}\left(G_{t:t+n} - Q_{t+n-1}(s_t, a_t)\right),
\end{equation}

\begin{equation}
    Q_{t+n}(s_t, a_t) \leftarrow Q_{t+n-1}(s_t, a_t) + \alpha  \rho_{t+1:t+n-1} \left(\sum_{k=t}^{t+n-1} \gamma^{k-t} \delta_{k} \right).
\end{equation}

where, $\delta_{t+k}$ is defined as per equations \ref{eqn:delta_eq1} and \ref{eqn:delta_eq2}. Note, we use $\rho_{t+1:t+n-1}$ and not $\rho_{t+1:t+n}$ as in any N-step expected SARSA such as this one, all possible actions are taken into account in the last state; the one
actually taken has no effect and does not have to be corrected for \citep[Page 150]{sutton2018reinforcement}. One can further write this recursively using per-decision importance sampling \citep{sutton2018reinforcement, precup2000eligibility}, but it is not essential to our derivations.

\textbf{N-step Soft Q-learning (off-policy with Tree Backup)}

We next present N-step Soft Q-learning using the Tree Backup algorithm. N-step soft Q-learning with importance sampling only uses the expectation over actions in the last time step. Tree Backup instead uses it at every step. This provides the following advantages: (1) reduces the variance due to the importance sampling ratio, (2) an importance sampling ratio does not need to be computed, thus the behavioural policy $b$ does not need to be stationary, Markov, or even known \citep{de2018multi, precup2000eligibility}. 

We begin by writing the N-step return under the Boltzmann policy after taking action $a_t$ in state $s_t$ in the Tree Backup format. Note, this is the soft-Bellman optimal return regardless of the behavioural policy which chooses actions $a_t, a_{t+1}, a_{t+2}, ...$ leading to states $s_{t+1}, s_{t+2}, s_{t+3}, ...$ respectively.

\begin{equation}
    G_{t:t+n} \doteq  r_{t+1} + \gamma V_{\pi^{\mathcal{B}}_Q}(s_{t+1})
    \label{eqn:startofTB}
\end{equation}

Using equation \ref{eqn:V_pi_relationship}, we get, 

\begin{equation}
    G_{t:t+n} \doteq  r_{t+1} + \gamma \left( \sum_{a} \pi^{\mathcal{B}}_Q (a|s_{t+1}) Q_t (s_{t+1}, a) - \tau \text{KL}_{t+1} \right)
\end{equation}

We can now write it in Tree-Backup format, 

\begin{equation}
\begin{split}
    G_{t:t+n} &\doteq r_{t+1} + \gamma \sum_{a\neq a_{t+1}} \pi^{\mathcal{B}}_Q (a|s_{t+1}) Q_t (s_{t+1}, a)  - \gamma \tau \text{KL}_{t+1}\\
    &+ \gamma  \pi^{\mathcal{B}}_Q(a_{t+1}|s_{t+1}) \left( r_{t+2} + \gamma \sum_{a\neq a_{t+2}} \pi^{\mathcal{B}}_Q (a|s_{t+2}) Q_{t+1} (s_{t+2}, a) - \gamma\tau \text{KL}_{t+2}\right) \\
    &+ \gamma^2 \pi^{\mathcal{B}}_Q(a_{t+2}|s_{t+2}) \pi^{\mathcal{B}}_Q(a_{t+1}|s_{t+1}) \left(r_{t+3} + \gamma \sum_{a\neq a_{t+3}} \pi^{\mathcal{B}}_Q (a|s_{t+3}) Q_{t+2} (s_{t+3}, a) - \gamma \tau \text{KL}_{t+3}\right) \\
    &+ ... \\
    &+ \gamma^{n-1} \prod_{i=t+1}^{min(t+n-1, T-1)} \pi^{\mathcal{B}}_Q(a_{i}|s_{i}) \left( r_{t+n} + \gamma \sum_a \pi^{\mathcal{B}}_Q (a|s_{t+n}) Q_{t+n-1} (s_{t+n}, a) - \gamma \tau \text{KL}_{t+n}\right)
\end{split}
\label{eqn:TB_format}
\end{equation}

This is visualised as follows: The update is from the estimated action values of the leaf nodes of the tree. The action nodes in the interior, corresponding to the actual actions taken, do not participate. Each leaf node contributes to the target with a weight proportional to its probability of occurring under the target policy. 

This can now be written recursively as,

\begin{equation}     
G_{t:t+n} \doteq r_{t+1} - \gamma \tau \text{KL}_{t+1} + \gamma \sum_{a\neq a_{t+1}} \pi^{\mathcal{B}}_Q (a|s_{t+1}) Q_t (s_{t+1}, a) + \gamma \pi^{\mathcal{B}}_Q(a_{t+1}|s_{t+1}) G_{t+1:t+n} 
\end{equation} 


Alternatively, it can also be compactly written in terms of temporal difference errors, by using the following relation from equation \ref{eqn:V_Q}: 

\begin{equation}
\begin{split}
    \sum_{a\neq a_{k}} \pi^{\mathcal{B}}_Q (a|s_{k}) Q_{k-1} (s_{k}, a) &=  \sum_{a} \pi^{\mathcal{B}}_Q (a|s_{k}) Q_{k-1} (s_{k}, a) - \pi^{\mathcal{B}}_Q (a_{k}|s_{k}) Q_{k-1} (s_{k}, a_{k}))\\
    &= V_Q(s_k) + \tau \text{KL}_{k} - \pi^{\mathcal{B}}_Q (a_{k}|s_{k}) Q_{k-1} (s_{k}, a_{k}))\\
\end{split}    
\end{equation}

By substituting this relation in equation \ref{eqn:TB_format}, we observe that the $\tau \text{KL}_k$ terms perfectly cancel out. Furthermore, at the $n$-step horizon (the leaf node), we simply terminate the expansion with the state-value $V_Q(s_{t+n})$. This allows us to write the Tree-Backup return in terms of TD-errors as follows: 

\begin{equation}
\begin{split}
    G_{t:t+n} &\doteq r_{t+1} + \gamma \left( V_Q(s_{t+1}) - \pi^{\mathcal{B}}_Q(a_{t+1}|s_{t+1}) Q_t(s_{t+1}, a_{t+1})\right)  \\
    &+ \gamma \pi^{\mathcal{B}}_Q(a_{t+1}|s_{t+1}) \left( r_{t+2} + \gamma V_Q(s_{t+2}) - \gamma \pi^{\mathcal{B}}_Q (a_{t+2}|s_{t+2}) Q_{t+1} (s_{t+2}, a_{t+2}) \right) \\
    &+ \gamma^2 \pi^{\mathcal{B}}_Q(a_{t+2}|s_{t+2}) \pi^{\mathcal{B}}_Q(a_{t+1}|s_{t+1}) \left(r_{t+3} + \gamma V_Q(s_{t+3}) - \gamma \pi^{\mathcal{B}}_Q (a_{t+3}|s_{t+3}) Q_{t+2} (s_{t+3}, a_{t+3}) \right) \\
    &+ ... \\
    &+ \gamma^{n-1} \left[\prod_{i=t+1}^{min(t+n-1, T-1)} 
    \pi^{\mathcal{B}}_Q(a_{i}|s_{i}) \right] \left( r_{t+n} + \gamma V_Q(s_{t+n}) \right)
\end{split}    
\end{equation}

If we combine $r_{k+1} + \gamma V_Q(s_{k+1})$ with the last term of the (previous) $k$-th term, and add and subtract $Q_{t-1}(s_t, a_t)$ for the first term, the sum telescopes. We define the Tree Backup TD-error for all $k \geq t$ as:

\begin{equation}
    \delta^{TB}_k = r_{k+1} + \gamma V_Q(s_{k+1}) - Q_{k-1}(s_k, a_k)
\end{equation}

This yields the exact $n$-step return:

\begin{equation}
    G_{t:t+n} = Q_{t-1}(s_t, a_t) + \sum_{k=t}^{min(T-1, t+n-1)} \left[\delta^{TB}_k \prod_{i=t+1}^{k} \gamma \pi^{\mathcal{B}}_Q(a_i|s_i) \right]
\end{equation}

Again, if we assume the approximate Q-values are unchanging (similar to Exercise 7.11 in \citet{sutton2018reinforcement}), this gives us our Q-update equation as follows:

\begin{equation}
    Q_{t+n}(s_t, a_t) \leftarrow Q_{t+n-1}(s_t, a_t) + \alpha \left(\sum_{k=t}^{min(T-1, t+n-1)} \delta^{TB}_k \prod_{i=t+1}^{k} \gamma \pi^{\mathcal{B}}_Q(a_i|s_i)  \right).
\end{equation}

These updates lead to the estimation of off-policy multi-step returns under any behavioural policy, without knowing the behavioural policy.

Note, that if one starts the Tree Backup derivation with $V_Q(s_t+1)$ instead of $V_{\pi^{\mathcal{B}}_Q} (s_{t+1})$, then this leads to an alternate equivalent derivation in terms of the default policy instead of the Boltzmann policy (which requires calculating TD-errors under the default policy as well). We think this alternate derivation is less relevant as the agent is the target policy for the agent is the soft-Bellman optimal Boltzmann policy; therefore, we focus on the derivation in terms of the Boltzmann policy.

This concludes our novel derivations of off-policy N-step extensions of Soft Q-learning, using either importance sampling or Tree-Backup. One may further aspire to unify these two multi-step off-policy methods, as done in the standard RL setting by \citet{de2018multi}, but it is not essential to the current work and is left as future work.

\subsection*{Appendix 2: Novel derivations extending N-step soft Q-learning to an elegant algorithm with eligibility traces}

\textbf{Soft Q($\lambda$) (on-policy with Boltzmann policy)}

Here, we build upon the N-step Soft Q-learning results to develop Soft Q($\lambda$), a solution using eligibility traces.

We define a $\lambda$-return, which is the weighted summation of $n$-step returns \citep{sutton2018reinforcement}.

\begin{equation}
    G_t^{\lambda} = (1-\lambda) \sum_{n=1}^{\infty} \lambda^{n-1} G_{t:t+n}
\end{equation}

To simplify the derivation, we define the  Boltzmann backup operator following \citet{schulman2017equivalence},

\begin{equation}
\begin{split}
    [\mathcal{T}_{\pi_{Q}^{\mathcal{B}}}Q](s,a) &= \mathbb{E}_{(s',r) \sim p(s',r|s,a)} \left[ r + \gamma \tau \log \mathbb{E}_{a' \sim \pi^d} [\exp(Q(s',a')/\tau) ] \right] \\
    &= \mathbb{E}_{(s',r) \sim p(s',r|s,a)} \left[ r + \gamma V_Q(s')] \right]\\
\end{split}
\end{equation}

We can now define the SARSA($\lambda$) version of this backup operator under the Boltzmann policy,  $[\mathcal{T_{\pi_{Q}^{\mathcal{B}},\lambda}Q}](s,a)$, as follows.

\begin{equation}
    G_t^{\lambda} = [\mathcal{T}_{\pi_{Q}^{\mathcal{B}},\lambda}Q]= (1-\lambda) ( 1 + \lambda \mathcal{T}_{\pi_{Q}^{\mathcal{B}}} + (\lambda \mathcal{T}_{\pi_{Q}^{\mathcal{B}}})^2 + ...) \mathcal{T}_{\pi_{Q}^{\mathcal{B}}}Q
\end{equation}

Based on n-step methods, we can derive it to be,
\begin{equation}
    G_t^{\lambda} = [\mathcal{T}_{\pi_{Q}^{\mathcal{B}},\lambda}Q](s,a) = Q(s,a) + \mathbb{E} \left[ \sum_{k=t}^{\infty} (\gamma \lambda) ^{k-t}  \delta_k \right]
\end{equation}

where,

\begin{equation}
    \delta_{k} = r_{k+1} - \tau \text{KL}_{k} + \gamma V_Q(s_{k+1}) - V_Q(s_{k})
    \label{eqn:softqlambda-onpolicy-TDerror}
\end{equation}

Note that this TD-error exactly recovers the soft Bellman error derived in Equation 43 of \citet{schulman2017equivalence}, confirming our foundational $n$-step derivation aligns with established on-policy methods. 

The update rule using $G_t^{\lambda}$, with a forward-view but offline algorithm is,

\begin{equation}
     Q_{t+1}(s,a) \leftarrow Q_{t}(s,a) + \alpha (G_t^{\lambda} - Q_t(s,a))
\end{equation}

This can be approximated using a backwards view (SARSA($\lambda$)-like) online algorithm under the Boltzmann policy, with eligibility traces ($e_t$) and the TD-errors as mentioned above in equation \ref{eqn:softqlambda-onpolicy-TDerror} ($\delta_t$).

\begin{equation}
    Q_{t+1}(s,a) \leftarrow Q_{t}(s,a) + \alpha \delta_t e_t (s,a) \; \; \forall s,a
\end{equation}

and eligibility traces are updated as follows (in the tabular setting),

\begin{equation}
    e_t(s, a) = 
    \begin{cases}
        \gamma \lambda e_{t-1}(s, a) + 1, & \text{if } (s, a) = (s_t, a_t), \\
        \gamma \lambda e_{t-1}(s, a), & \text{otherwise},
    \end{cases}
\end{equation}

\textbf{Soft Q($\lambda$) (off-policy with Tree Backup)}

We next extend the algorithm to a full off-policy algorithm, developing upon the n-step method using the Tree Backup algorithm.

\begin{equation}
    G_t^{\lambda} \approx Q(s,a) +  \left[ \sum_{k=t}^{\infty}\delta^{TB}_k  \prod_{i=t+1}^{k}\gamma  \lambda \pi^{\mathcal{B}}_Q (a_i|s_i) \right]
\end{equation}

Which gives us an online off-policy soft Q($\lambda$) algorithm, similar to the previous one, but the eligibility trace update is adjusted with the target policy $\pi^{\mathcal{B}}_Q$,

\begin{equation}
    Q_{t+1}(s,a) \leftarrow Q_{t}(s,a) + \alpha \delta^{TB}_t e_t (s,a) \; \; \forall s,a
\end{equation}

where,

\begin{equation}
    \delta^{TB}_{t} = r_{t+1} + \gamma V_Q(s_{t+1}) - Q_t(s_{t}, a_{t})
    \label{eqn:softqlambda-offpolicy-TDerror}
\end{equation}

and,

\begin{equation}
    e_t(s, a) = 
    \begin{cases}
        \gamma \lambda \pi^{\mathcal{B}}_Q (a_t|s_t)  e_{t-1}(s, a) + 1, & \text{if } (s, a) = (s_t, a_t), \\
        \gamma \lambda \pi^{\mathcal{B}}_Q (a_t|s_t) e_{t-1}(s, a), & \text{otherwise},
    \end{cases}
\end{equation}

This concludes our derivation of a basic online off-policy Soft Q($\lambda$) algorithm. As is standard in temporal difference learning, this backward-view online algorithm serves as a computationally efficient approximation of the forward-view $\lambda$-return. Exact equivalence between the forward and backward views can be achieved by extending this framework to a "True Online" formulation (e.g., using Dutch traces). Future work can readily extend these derivations to include (1) function approximation, (2) true online equivalence, and (3) variance-reduction mechanisms such as V-trace, following the foundations laid out in Chapter 12 of \citet{sutton2018reinforcement}.

\end{document}